\documentclass[a4paper,fleqn]{cas-sc}

\usepackage[authoryear,longnamesfirst]{natbib}

\usepackage{algorithm}
\usepackage{algpseudocode}
\def\tsc#1{\csdef{#1}{\textsc{\lowercase{#1}}\xspace}}
\tsc{WGM}
\tsc{QE}

\begin{document}
\let\WriteBookmarks\relax
\def\floatpagepagefraction{1}
\def\textpagefraction{.001}

\shorttitle{Exploiting Causality Signals in Medical Images}    

\shortauthors{Carloni et al.}  

\title [mode = title]{Exploiting Causality Signals in Medical Images: A Pilot Study with Empirical Results}  



%

\author[1,2]{Gianluca Carloni}[orcid=0000-0002-5774-361X]

\cormark[1]
\cortext[1]{Corresponding author}


\ead{gianluca.carloni@isti.cnr.it}


\credit{Conceptualization, Methodology, Software, Validation, Formal analysis, Investigation, Data Curation, Writing - Original Draft, Writing - Review \& Editing, Visualization}

\affiliation[1]{organization={National Research Council, Institute of Information Science and Technologies (ISTI) "Alessandro Faedo"},
            addressline={Via Moruzzi, 1}, 
            city={Pisa},
            postcode={56127}, 
            state={Italy},
            country={IT}}

\affiliation[2]{organization={Department of Information Engineering, University of Pisa},
            addressline={Via Caruso, 16}, 
            city={Pisa},
            postcode={56122}, 
            state={Italy},
            country={IT}}

\author[1]{Sara Colantonio}[orcid=0000-0003-2022-0804]
\ead{sara.colantonio@isti.cnr.it}
\credit{Resources, Writing - Review \& Editing, Supervision, Project administration, Funding acquisition}

\begin{abstract}
We present a novel technique to discover and exploit weak causal signals directly from images via neural networks for classification purposes. This way, we model how the presence of a feature in one part of the image affects the appearance of another feature in a different part of the image. Our method consists of a convolutional neural network backbone and a causality-factors extractor module, which computes weights to enhance each feature map according to its causal influence in the scene. We develop different architecture variants and empirically evaluate all the models on two public datasets of prostate MRI images and breast histopathology slides for cancer diagnosis. We study the effectiveness of our module both in fully-supervised and few-shot learning, we assess its addition to existing attention-based solutions, we conduct ablation studies, and investigate the explainability of our models via class activation maps. Our findings show that our lightweight block extracts meaningful information and improves the overall classification, together with producing more robust predictions that focus on relevant parts of the image. That is crucial in medical imaging, where accurate and reliable classifications are essential for effective diagnosis and treatment planning.
\end{abstract}

\maketitle

\section{Introduction}\label{sec:introduction}
Automatic diagnosis models from medical data can potentially transform how patients are treated, especially in oncology. They could reduce the need for invasive tests and increase the likelihood of successful outcomes for the most severe cases.
In this regard, some successful examples of machine learning (ML) systems for medical diagnosis exist, such as colon cancer diagnosis from gene expression profiling data \citep{su2022colon}, diagnosis of neurological diseases using voice data \citep{wroge2018parkinson}, and identification of patients with pulmonary
hypertension using electronic health records \citep{kogan2023machine}.

In recent years, the concepts of causal inference and causal reasoning have received increasing attention across the Artificial Intelligence (AI) community. This trend began with the very first work of the computer scientist Judea Pearl on Bayesian networks and the mathematical formalization of causality, which enabled the creation of computational systems that can automatically model causality  \citep{pearl1985bayesian, pearl2009causality, pearl2018book}. Today, we have inspiring examples of the integration of causality into the ML community \citep{luo2020causal, scholkopf2022causality} and the deep learning research \citep{berrevoets2023causal}, with extensions to causal representation learning \citep{scholkopf2021toward}, causal discovery under distribution shifts \citep{perry2022causal} and with incomplete data \citep{wang2020causal}. Unfortunately, this line of research has always had in common the fact that the processed data are tabular, structured, not always real but simulated, and very often accompanied by a priori information about the process that generated them.

Unlike tabular data, when it comes to images, their representation does not include any explicit indications regarding objects or patterns. Instead, individual pixels are used to convey a particular scene visually, and image datasets do not usually provide labels describing the objects’ dispositions. Additionally, unlike video frames, a single image cannot reveal the dynamics of the appearance and change of objects in a scene. These critical issues could explain why images have been neglected by research on the tabular causal discovery, where instead, there are established algorithms \citep{spirtes1991algorithm,spirtes2000causation,chickering2002optimal}. 
A particular case would be discovering hidden causalities among objects in an image dataset, as suggested by \cite{terziyan2023causality}, who conceive a way to compute possible causal relationships within images. Although the idea is compelling, that work is preliminary, and a thorough investigation of the effectiveness of their method is lacking. 

In our work, we intervene in this lack and propose a way to discover and exploit weak causal signals within images without requiring prior knowledge and use them to enhance convolutional neural network (CNN) classifiers. By combining a regular CNN with the proposed causality-factors extraction module, we present a new scheme based on feature map enhancement to enable “causality-driven” CNNs. This way, we weight each feature map according to its causal influence in the scene, in an attention-inspired fashion.
We frame our system as an automatic diagnosis model from medical images since we study the efficacy of the proposed methods with extensive empirical evaluations on publicly available datasets of MRI images and histopathology slides. We study the effectiveness of our module both in fully-supervised and few-shot learning regimes, investigate its integration with existing attention-based solutions, and conduct ablation studies.
Besides investigating the quantitative aspect, we also explored the concept of explainability of AI in our evaluation. The results on class activation maps demonstrate that our method improves classification and produces more robust predictions by focusing on the relevant parts of the image, thus enhancing reliability, trustworthiness, and user confidence.

Our paper is structured as follows. First, in Sec. \ref{sec:preliminaries}, we provide the concepts behind the causal signals' interpretation in images. Then, we start Sec. \ref{sec:methods} by describing the novelty of our work, namely the methodological framework and the causality-factors extractor module we introduced. We also illustrate the datasets, the training scheme, and the evaluation details. Later, we present our main results in Sec. \ref{sec:results}, explore the significance of our findings in the general discussion in Sec. \ref{sec:discussion}, and pull the threads in Sec. \ref{sec:conclusions}.

\section{Causality signals in images}\label{sec:preliminaries}
\cite{lopez2017discovering} propose the idea of “causal disposition” as a simple way to understand the hidden causes in images instead of using the methods of do-calculus and causal graphs from Pearl’s framework \citep{pearl2009causality,pearl2018book}. 
In their view, by counting the number $C(A,B)$ of images in which the causal dispositions of artifacts $A$ and $B$ are such that $B$ disappears if one removes $A$, one can assume that artifact $A$ causes the presence of artifact $B$ when $C(A,B)$ is greater than the converse $C(B,A)$. For instance, they argue that the presence of a car causes the presence of a wheel, but not the other way around, because removing the car would make the wheel disappear, but removing the wheel would not make the car disappear. By studying such asymmetries, the authors find the causal direction between pairs of random variables representing features of objects and their contexts in images. Although the causal disposition concept is more primitive than the interventional approach, it could be the only way to proceed with limited a priori information.
This concept leads to the intuition that any causal disposition induces a set of asymmetric causal relationships between the artifacts from an image (features, object categories, etc.) that represent (weak) causality signals regarding the real-world scene. A point of contact with machine vision systems would be to automatically infer such asymmetries from an observed image dataset.

\begin{figure}
	\centering
		\includegraphics[width=0.99\textwidth]{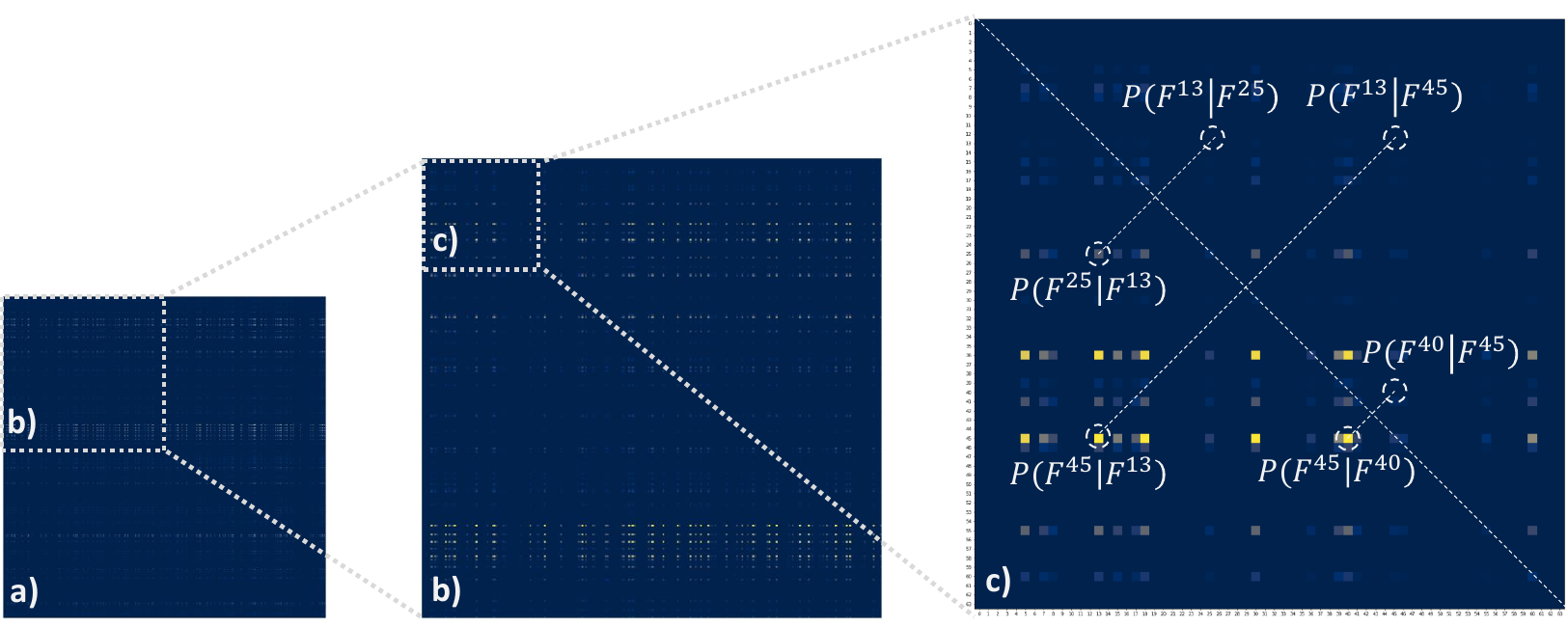}
	  \caption{Zoomed-in visualizations of a sample causality map computed with Eq. \ref{eq:causality_method_max} on $512$ feature maps extracted from an input image. (a) $512 \times 512$ original causality map; (b) $256 \times 256$ zoom-in of (a); (c) $64 \times 64$ zoom-in of (b), where dashed circles indicate exemplar elements and their corresponding elements opposite the main diagonal, representing conditional asymmetries of the type $P(F^i|F^j) \neq P(F^j|F^i)$. We can see that, for instance, $P(F^{25}|F^{13}) > P(F^{13}|F^{25})$, that is $F^{25} \rightarrow F^{13}$, and $P(F^{45}|F^{40}) > P(F^{40}|F^{45})$, that is $F^{45} \rightarrow F^{40}$.}
   \label{fig_cmapvisual}
\end{figure}

\cite{terziyan2023causality} suggest a way to compute estimates for possible causal relationships within images via CNNs.
CNNs obtain the essential features required for classification not directly from the pixel representation of the input image but through a series of convolution and pooling operations designed to capture meaningful features from the image. Convolution layers are responsible for summarizing the presence of specific features in the image and generating a set of feature maps accordingly. Pooling consolidates the presence of particular features within groups of neighboring pixels in square-shaped sub-regions of the feature map.
When a feature map $F^i$ contains only non-negative numbers (e.g., thanks to ReLU functions) and is normalized in the interval $[0,1]$, we can interpret its values as probabilities of that feature to be present in a specific location. For instance, $F^i_{r,c}$ is the probability that the feature $i$ is recognized at coordinates ${r,c}$.
By assuming that the last convolutional layer outputs and localizes to some extent the object-like features, we may modify the architecture of a CNN such that the $n \times n$ feature maps ($F^1,F^2,\dots F^k$) obtained from that layer got fed into a new module that computes pairwise conditional probabilities of the feature maps. The resulting $k \times k$ map would represent the causality estimates for the features and be called \textit{causality map}. 
Given a pair of feature maps $F^i$ and $F^j$ and the formulation that connects conditional probability with joint probability, $P(F^i|F^j) = \frac{P(F^i,F^j)}{P(F^j)}$, \cite{terziyan2023causality} suggest to heuristically estimate this quantity by adopting two possible methods, namely \textit{Max} and \textit{Lehmer}.
The \textit{Max} method considers the joint probability to be the maximal presence of both features in the image (each one in its location):
\begin{equation}
    P(F^i|F^j) = \frac{(\max_{r,c} F^i_{r,c})\cdot (\max_{r,c} F^j_{r,c})}{\sum_{r,c} F^j_{r,c}}
    \label{eq:causality_method_max}
\end{equation}
On the other hand, the \textit{Lehmer} method entails computing 
\begin{equation}
    P(F^i|F^j)_p = \frac{LM_p(F^i \times F^j)}{LM_p(F^j)}
    \label{eq:causality_method_lehmer}
\end{equation}
where $F^i \times F^j$ is a vector of $n^4$ pairwise multiplications between each element of the two $n \times n$ feature maps, while $LM_p$ is the generalized Lehmer mean function \citep{bullen2003handbook} with parameter $p$, which is an alternative to power means for interpolating between minimum and maximum of a vector $x$ via harmonic mean ($p=-1$), arithmetic mean ($p=0$), and contraharmonic mean ($p=1$):
$LM_p(x) = \frac{\sum_{k=1}^n x_k^{p+1}}{\sum_{k=1}^n x_k^p}$.
Equations \ref{eq:causality_method_max} and \ref{eq:causality_method_lehmer} could be used to estimate asymmetric causal relationships between features $F^i$ and $F^j$, since, in general, $P(F^i|F^j) \neq P(F^j|F^i)$. By computing these quantities for every pair $i$ and $j$ of the $k$ feature maps, the $k \times k$ causality map is obtained. We interpret asymmetries in such probability estimates as weak causality signals between features, as they provide some information on the cause-effect of the appearance of a feature in one place of the image, given the presence of another feature within some other places of the image. Accordingly, a feature may be deemed to be the reason for another feature when $P(F^i|F^j) > P(F^j|F^i)$, that is ($F^i \rightarrow F^j$), and vice versa.  
As an example, Figure \ref{fig_cmapvisual} depicts a causality map to give a visual interpretation of this concept.  

In this work, we integrate a regular CNN with a new causality-extraction module to explore the features and causal relationships between them extracted during training. The previous work that inspired us \citep{terziyan2023causality} is preliminary, and we introduce a novel attention-like scheme based on feature maps enhancement to enable “causality-driven” CNNs, providing an extensive empirical evaluation of the impact of this new introduction on real data.
We hypothesize that it would be possible and reasonable to get some weak causality signals from the individual images of some medical datasets without adding primary expert knowledge and leverage them to better guide the learning phase. 
Ultimately, a model trained in such a manner would exploit weak causal dispositions of objects in the image scene to distinguish the tumor status of a medical image.

\section{Material and Methods}\label{sec:methods}
\subsection{Embedding causality into CNNs}
\label{sec:embedding_causality}
\begin{figure}
	\centering
		\includegraphics[width=0.99\textwidth]{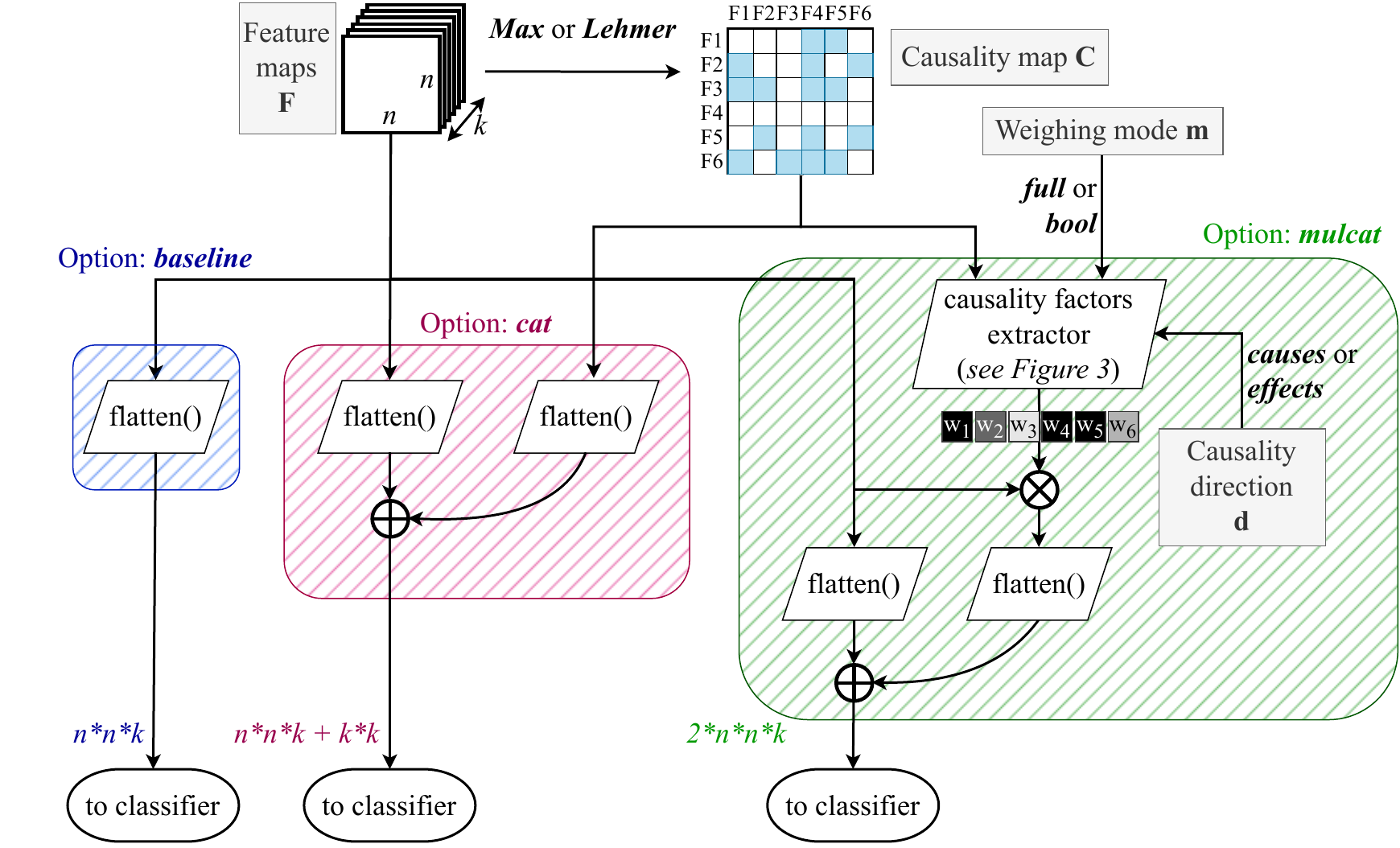}
	  \caption{Overview of the different settings investigated in this work: \textit{Baseline}, \textit{Cat}, and the proposed \textit{Mulcat}. Assuming $k=6$ feature maps as an example, the tensor \textbf{F} of feature maps that are obtained from a CNN just before the classifier can be either flattened and used as they are (Option \textit{baseline}) or can be leveraged to compute the causality map \textbf{C} via the \textit{Max} or \textit{Lehmer} method. Once obtained, \textbf{C} can be flattened as well and concatenated to the feature maps (Option \textit{Cat}) or fed to our proposed causality factors extractor (see Figure \ref{fig:caufacextractor}) to implement the Option \textit{Mulcat}. The latter produces a vector of causality factors that weighs the feature maps obtaining a causality-driven version of them, which is then concatenated to the original ones and fed to the classifier. Weighing mode \textbf{m} and causality direction \textbf{d} are two external signals used to tune the functioning of the system. This image is best seen in color.}
   \label{fig:overview}
\end{figure}

Usually, a CNN performs image classification based on the final set of (flattened) $n\times n \times k$ feature maps obtained just before the dense layers that constitute the classifier. In the following, we describe how the architecture of such a regular CNN (baseline) might be modified to make the classifier consider the information entailed in the estimated causality map.

Feature concatenation is a basic (yet popular) way to embed additional information in CNNs. Indeed, by concatenating the flattened causality map to the flattened set of feature maps just before the classifier, \cite{terziyan2023causality} let the CNN learn how these causality estimates influence image classification. That means that in addition to the $n\times n \times k$ features, the fully connected layers of the classifier will now have a $k \times k$ input, and the weights for the corresponding connections (i.e., actual causality influences) will be learned by back-propagation the same way as other neural network parameters. We will call this method the \textbf{Cat} (\textit{con\textbf{cat}enate}) option (see the magenta box in Figure \ref{fig:overview}). 

Alternatively, one could enhance or penalize parts of the existing information according to the newly gained one. Our proposition here is a new way to exploit the causality map: this time, it is used to compute a vector of causality factors that multiply (i.e., weighs) the feature maps so that each feature map is strengthened according to its causal influence within the image's scene. 
After multiplication, the obtained causality-driven version of the feature maps is flattened and concatenated to the flattened original ones, producing a $2 \times n\times n \times k$ input to the classifier. We will call this method the \textbf{Mulcat} (\textit{\textbf{mul}tiply and con\textbf{cat}enate}) option (see the green box in Figure \ref{fig:overview}).

\begin{figure}
	\centering
		\includegraphics[width=0.99\textwidth]{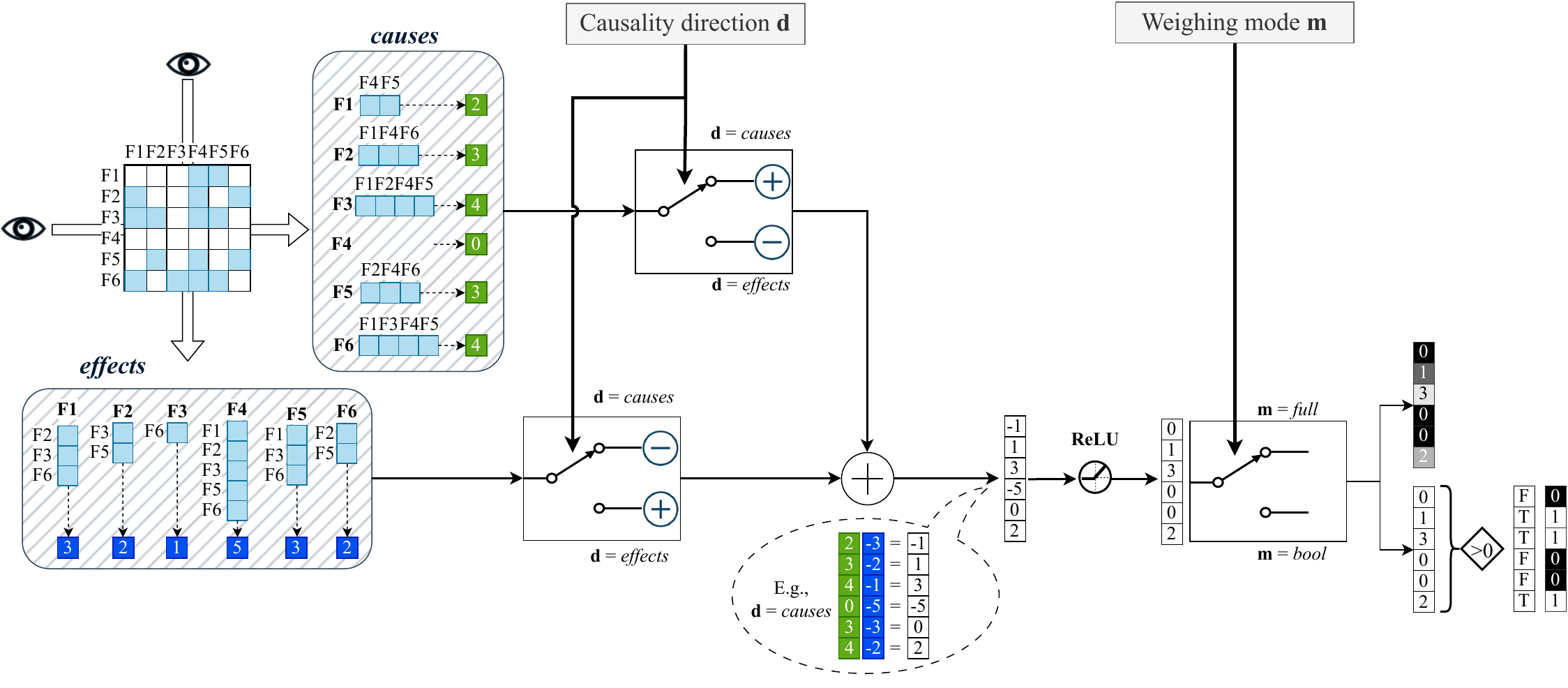}
	  \caption{The internals of the proposed \textit{causality factors extractor} block of Figure \ref{fig:overview} given an example causality map. Cyan squares in the causality map indicate whether the probability value of one element is greater than its element opposite the main diagonal. The \textit{causes} box shows how the causality map is processed row-wise for each feature map: the number of times that feature is a cause of another feature is registered. Similarly, the \textit{effects} box shows how the causality map is processed column-wise for each feature map. Before being summed element-wise, those two vectors are either passed as they are or the sign of their elements is reversed according to the causality direction \textbf{d}. The obtained vector is rectified and then returned as it is or passed through boolean filtering depending on the weighing mode \textbf{m}. This image is best seen in color.}
   \label{fig:caufacextractor}
\end{figure}

At the core of the \textbf{Mulcat} option stands our \textit{causality factors extractor} module, which yields the vector of weights needed to multiply the feature maps (see Figure \ref{fig:caufacextractor}).
The main idea here is to look for asymmetries between elements opposite the main diagonal of the causality map, as they represent conditional asymmetries entailing possible cause-effect relationships (e.g., Figure \ref{fig_cmapvisual}). Indeed, some features may be more often found on the left side of the arrow (i.e., $F\rightarrow$) than on the right side (i.e., $\rightarrow F$). 
Accordingly, the $2$D causality map is processed row-wise and column-wise. In the former case, we register the number of times each feature map $F^i$ was found to cause another feature map $F^j$, that is, $P(F^i|F^j) > P(F^j|F^i)$. This way, we obtain a vector of values that quantify how much those feature maps can be called "\textit{causes}." Conversely, in the column-wise processing, we register the number of times each feature map $F^j$ was found to be caused by another feature map $F^i$, obtaining a vector of values that quantify how much the feature maps can be deemed "\textit{effects}."

At this point, we propose two variants to the model's functioning. We allow an external signal $\textbf{d}$ to represent the causality direction of analysis, which can be either \textit{causes} or \textit{effects}. When \textbf{d} $=$ \textit{causes}, the vector of \textit{causes} (obtained row-wise) is not altered, while the sign is changed to the elements of the \textit{effects} vector (obtained column-wise). Hence, as those two vectors enter a summation point, the difference between \textit{causes} and \textit{effects} is obtained as the weight vector. On the other hand, when \textbf{d} $=$ \textit{effects}, the vector of \textit{effects} is not altered, while it is to the vector of \textit{causes} that the sign is changed. Therefore, the difference between \textit{effects} and \textit{causes} is obtained at the summation point. As a result, the obtained weight vector is rectified to set any negative elements to zero.

In addition, we conceive two variants of the model controlled by another external signal \textbf{m}, that represents the weighing mode and can be one of: 

\begin{itemize}
    \item \textbf{full}. The vector of non-negative causality factors is left at its full count, being returned as it is. As a result of this choice, the model weighs features more according to their causal importance (a feature that is \textit{cause} $10$ times more than another receives $10$ times more weight).

    \item \textbf{bool}. The factors undergo boolean thresholding where all the non-zero factors are assigned a new weight of $1$ and $0$ otherwise. As a result, this choice is more conservative and assigns all features that are most often \textit{causes} the same weight.
\end{itemize}
In the following sections, we describe the data used for our empirical evaluations, the different types of model architectures we utilized, and the implementation details of the training process.

\subsection{Datasets}
To validate our proposed methods, we utilized multiple publicly available medical imaging datasets. To begin with, we exploited the Breast cancer Histopathological Image (BreakHis) dataset \citep{Spanhol15}. On the other hand, we used the dataset from the PI-CAI challenge \citep{Saha23}, comprising multi-parametric MRI (mpMRI) acquisitions of the prostate. 

\subsubsection{BreakHis dataset}
The dataset has 7909 microscopic images of breast tumor tissues aggregated from 82 subjects at magnification levels of 40, 100, 200, and 400. There are eight classes in this dataset, namely adenosis, tubular adenoma, fibroadenoma, phyllodes tumor, papillary carcinoma, lobular carcinoma, mucinous carcinoma, and ductal carcinoma. In addition, a binary classification was provided, namely, \textit{benign} and \textit{malignant} lesions. In particular, the first four classes represent benign lesions, while the last four represent malignant lesions. We considered the images with a magnification level of 400 for a total of 1819 images. We split this dataset into training (1235 images), validation (218 images), and test (366 images) sets, ensuring class balancing according to the binary classification, i.e., benign and malignant. 

\begin{figure}
	\centering
		\includegraphics[width=0.99\textwidth]{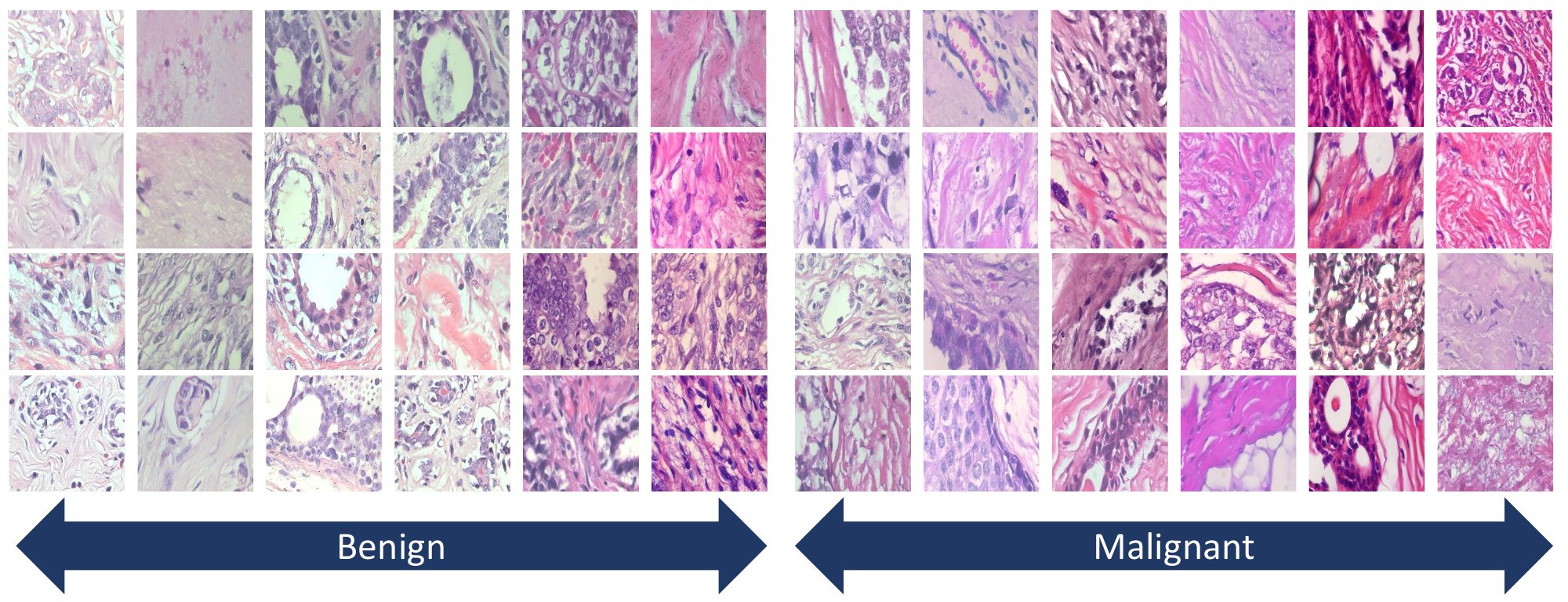}
	  \caption{Some benign and malignant samples from the utilized BreakHis dataset. This image is best seen in color.}
   \label{fig:dataset_breakhis}
\end{figure}

In this study, we utilize the processed version of the BreakHis dataset, curated by \cite{Pereira23} to be used in ML tasks. Indeed, the original images were resized to 224x224 pixels and organized according to binary and multiclass classification tasks. We further resize the images to a consistent 128x128 pixel matrix. Some samples from the utilized dataset are presented in Figure \ref{fig:dataset_breakhis}.

\subsubsection{PI-CAI dataset}
From the available $1500$ acquisitions, we only selected  T2-weighted (T2w) images. Within this cohort of patients and respective scans, some cases didn't have any tumors (i.e., they had no biopsy examination), while others had cancer lesions. For each of the latter, the dataset contained biopsy reports expressing the severity as Gleason Score (GS). In anatomopathology, a GS of $1$ to $5$ is assigned to the two most common patterns in the biopsy specimen based on the cancer severity. The two grades are then added together to determine the GS, which can assume all the combinations of scores from "$1$+$1$" to "$5$+$5$". Additionally, the dataset included the assigned GS's group affiliation, defined by the International Society of Urological Pathology (ISUP) \citep{Egevad16}, ranging from $1$ to $5$, which provides the tumor severity information at a higher granularity level.
In this study, we included both cancerous and no-tumor patients. From the former case, we only considered lesions with GS $\geq 3+4$ (ISUP $\geq 2$) and selected only the slices containing lesions by exploiting the expert annotations of the disease provided in the dataset. For the latter case, we considered all the available slices. In the end, we obtained a total number of $4159$ images (from $545$ patients), with a balanced distribution over the two classes: $2079$ tumor images vs. $2080$ no-tumor images.
To constitute our subsets, we divided the available images into training ($2830$), validation ($515$), and testing ($814$) subsets. During the splitting process, we ensured patient stratification (i.e., images of the same patient were grouped to prevent data leakage) and class balancing.

\begin{figure}
	\centering
		\includegraphics[width=0.99\textwidth]{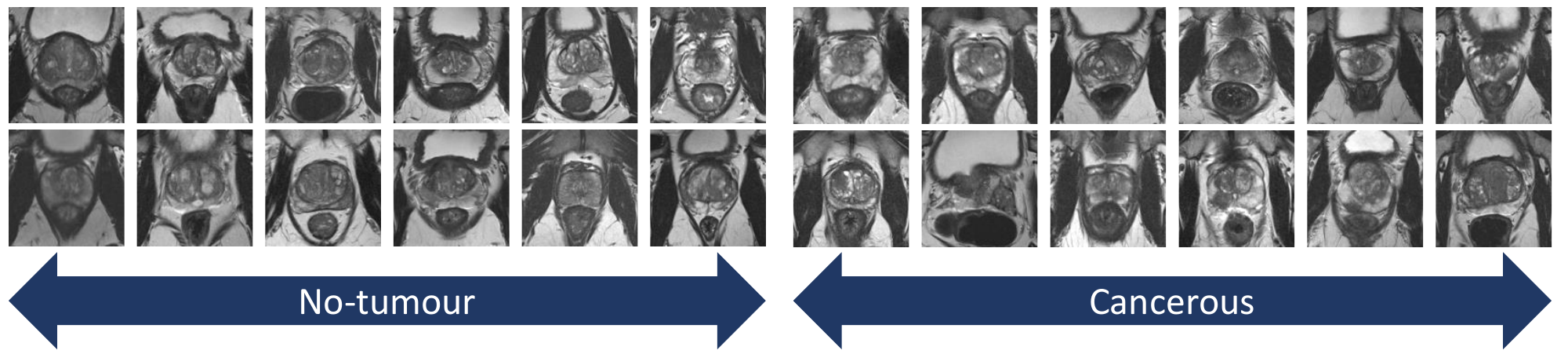}
	  \caption{Some no-tumour and cancerous samples from the utilized PI-CAI dataset.}
   \label{fig:dataset_picai}
\end{figure}

We utilized the provided whole prostate segmentation to extract the mask centroid for each slice. We then standardized the field of view (FOV) at $100$ mm in both $x$ ($FOV_x$) and $y$ ($FOV_y$) directions to ensure consistency across all acquisitions and subsequently cropped each image based on this value around its centroid. To determine the number of rows ($N_{rows}$) and columns ($N_{cols}$) corresponding to the fixed FOV, we utilized the pixel spacing in millimeters along the $x$-axis ($px$) and the $y$-axis ($py$). The relationships used to derive the number of columns and rows are $N_{cols} = \frac{FOV_x}{px}$ and $N_{rows} = \frac{FOV_y}{py}$, respectively.
Furthermore, we resized all the images to a uniform matrix size of $96 \times 96$ pixels to maintain consistent pixel counts. Finally, we performed image normalization using an in-volume method. That involved calculating the mean and standard deviation of all pixels within the volume acquisition and normalizing each image based on these values using the z-score technique. Some samples from the utilized dataset are presented in Figure \ref{fig:dataset_picai}.

\subsection{Architecture and training}
For each dataset, we built different CNN models to automatically classify input images in the two classes according to their diagnosis labels under full supervision. As for the architectures, we used the popular ResNet18 as the backbone for all the causality-driven models.
To handle images of different sizes in image recognition, many common architectures use an adaptive average pooling layer that outputs a $1\times1$ shape before the classifier. It does this by adjusting its parameters (such as kernel size, stride, and padding) based on the input size. However, this reduces the dimensionality of the feature maps and ignores their 2D structure, which is needed for finding causalities. Therefore, we replaced the \textit{AdaptiveAvgPool2D} layer of the ResNet18 with an identity layer in our experiments.

As described in Section \ref{sec:embedding_causality}, we could integrate the information of the causality map into the CNN classification in different manners. In this work, we developed six types of models for each  dataset and trained them to test the efficacy of the newly proposed \textit{Mulcat} architectures on medical image classification, namely:
\begin{itemize}
    \item \textbf{ResNet18}. This model is a regular ResNet18 architecture to serve as a baseline, where we replaced its \textit{AdaptiveAvgPool2D} layer with an identity layer. See Figure \ref{fig:overview} (blue box) for a visual representation.
    \item \textbf{ResNet18 $+$ Cat}. This is a ResNet18 model we modified to embed the causal information via concatenation as in \cite{terziyan2023causality}. See Figure \ref{fig:overview} (magenta box) for a visual representation. 
    \item \textbf{ResNet18 $+$ Mulcat (full, causes)}. This variant exploits our \textit{causality factors extractor} to obtain weights for the feature maps. In this model, we set the causality direction \textbf{d} = \textit{causes} and the weighing mode \textbf{m} = \textit{full}.
    \item \textbf{ResNet18 $+$ Mulcat (bool, causes)}. It is similar to the previous, but we set the weighing mode to \textbf{m} = \textit{bool}.
    \item \textbf{ResNet18 $+$ Mulcat (full, effects)}. This variant turns the way the set of causality factors is obtained within our \textit{causality factors extractor} by setting the causality direction \textbf{d} = \textit{causes}. We use \textbf{m} = \textit{full} in this model.
    \item \textbf{ResNet18 $+$ Mulcat (bool, effects)}. It is analogous to the previous, but setting the weighing mode to \textbf{m} = \textit{bool}.
\end{itemize}

As shown in Figure \ref{fig:overview}, the different types of models we investigated expose the classifier to a different number of input features. Therefore, the classifier is modified for each type according to the number of new neurons entering the fully-connected layer.

We optimized the way we computed the causality maps (using either the \textit{Max} option (Eq. \ref{eq:causality_method_max}) or the \textit{Lehmer} (Eq. \ref{eq:causality_method_lehmer})) and, for the \textit{Lehmer} option, we tried six different values of its parameter \textit{p}: [$-100, -2, -1, 0, 1, 100$]. Consequently, for each dataset, we trained seven models for each of the five types of causality-driven models, resulting in $35$ causality-driven models plus one baseline model. 
%
We provide the pseudo-code for the algorithms utilized to compute the causality maps and the proposed causality factors in Algorithm \ref{alg:cmap} and Algorithm \ref{alg:cfactors}, respectively. 

\begin{algorithm}
\caption{Causality map computation}\label{alg:cmap}
\begin{algorithmic}[1]
\Require $\textbf{x}\gets$ feature maps [$k$,$n$,$n$], $CM\gets$ computation method, $lehmer\_p \gets$ Lehmer mean power.
\State $\textbf{x} \gets \textbf{x}/max(\textbf{x})$ \Comment{Normalize in range [$0-1$] by dividing for maximum activation across all maps}
\State $\textbf{cmap} \gets \textbf{0}$ \Comment{Size of cmap: [$k$,$k$]}
\If {$CM == $ \textit{Max}} \Comment{Eq. \ref{eq:causality_method_max}}
    \State $\textbf{sumValues} \gets sum(flatten(\textbf{x}),dim=1)$ \Comment{Compute sum of values for each feature}
    \State $\textbf{maxValues} \gets max(flatten(\textbf{x}),dim=1)$ \Comment{Get maximum values of each feature}
    \State $\textbf{prod} \gets outerProduct(\textbf{maxValues},\textbf{maxValues})$ \Comment{Numerator of Eq. \ref{eq:causality_method_max} for each $i$,$j$}
    \State $\textbf{cmap} \gets \textbf{prod}/\textbf{sumValues}$
\Else \If {$CM ==$ \textit{Lehmer}} \Comment{Eq. \ref{eq:causality_method_lehmer}}
    \State $\textbf{x}\gets flatten(\textbf{x})$
    \State $\textbf{crossMat} \gets outerProduct(\textbf{x},\textbf{x})$ \Comment{Pairwise multiplications between each element of the features}
    \State $\textbf{num\_a} \gets sum(exp(\textbf{crossMat},lehmer\_p+1))$
    \State $\textbf{num\_b} \gets sum(exp(\textbf{crossMat},lehmer\_p))$
    \State $\textbf{num} \gets \textbf{num\_a}/\textbf{num\_b}$ \Comment{Numerator of Eq. \ref{eq:causality_method_lehmer}}
    \State $\textbf{den\_a} \gets sum(exp(\textbf{x},lehmer\_p+1))$
    \State $\textbf{den\_b} \gets sum(exp(\textbf{x},lehmer\_p))$
    \State $\textbf{den} \gets \textbf{den\_a}/\textbf{den\_b}$ \Comment{Denominator of Eq. \ref{eq:causality_method_lehmer}}
    \State $\textbf{cmap} \gets \textbf{num}/\textbf{den}$
    \EndIf
\EndIf
\State \textbf{return cmap}
\end{algorithmic}
\end{algorithm}

\begin{algorithm}
\caption{Causality factors extraction}\label{alg:cfactors}
\begin{algorithmic}[1]
\Require $\textbf{x}\gets$ feature maps [$k$,$n$,$n$], $\textbf{cmap}\gets$ causality map [$k$,$k$], $d\gets$ causality direction, $m\gets$ weighing mode.
\State $\textbf{triu} \gets triu(\textbf{cmap},diag=1)$ \Comment{Upper triangular matrix}
\State $\textbf{tril} \gets tril(\textbf{cmap},diag=-1).transpose()$ \Comment{Lower triangular matrix}
\State $\textbf{bool\_ij} = (tril>triu).transpose()$
\State $\textbf{bool\_ji} = (triu>tril)$
\State $\textbf{bool\_matrix} = \textbf{bool\_ij} + \textbf{bool\_ji}$ \Comment{Sum of booleans is the OR logic}
\State $\textbf{by\_col} = sum(\textbf{bool\_matrix}, dim=1)$ \Comment{Obtain the \textit{causes} view}
\State $\textbf{by\_row} = sum(\textbf{bool\_matrix}, dim=0)$ \Comment{Obtain the \textit{effects} view}
\If {$d ==$ \textit{causes}}
    \State $\textbf{mul\_factors} = ReLU(\textbf{by\_col} - \textbf{by\_row})$ \Comment{Difference between \textit{causes} and \textit{effects}}
    \If {$m ==$ \textit{full}}
        \State \textbf{return mul\_factors}
    \Else \If {$m ==$ \textit{bool}}
        \State $\textbf{mul\_factors} = 1*(\textbf{mul\_factors}>0)$ \Comment{Boolean thresholding}
        \State \textbf{return mul\_factors}
    \EndIf
    \EndIf
\Else \If {$d ==$ \textit{effects}}
    \State $\textbf{mul\_factors} = ReLU(\textbf{by\_row} - \textbf{by\_col})$ \Comment{Difference between \textit{effects} and \textit{causes}}
    \If {$m ==$ \textit{full}}
        \State \textbf{return mul\_factors}
    \Else \If {$m ==$ \textit{bool}}
        \State $\textbf{mul\_factors} = 1*(\textbf{mul\_factors}>0)$ \Comment{Boolean thresholding}
        \State \textbf{return mul\_factors}
\EndIf
\EndIf
\EndIf
\EndIf
\end{algorithmic}
\end{algorithm}

Regarding the training phase, we utilized the cross-entropy loss as the criterion and Adam as the optimizer, as well as performed data augmentation (random horizontal flip) at training time. We trained the models for $200$ epochs and set up a learning rate (LR) scheduler to decrease the LR during training. Specifically, the scheduler starts by multiplying the LR by $1.0$ after the first epoch, and then this factor linearly decreases to $0.1$ at epoch $200$.
As for models' hyperparameters, we investigated different values of initial LR ($0.01$ and $0.001$) and of weight decay ($0.01$, $0.001$, and $0.0001$). Accordingly, for each dataset, we trained the $36$ models for each of the six combinations of hyperparameters and chose the best-performing model on the validation set.
To prevent our results from being biased due to the random processes of the algorithms, we repeated the entire analysis ($216$ experiments) four times with different starting seeds that govern the random processes of the scripts.

\subsection{Quantitative evaluation}
During training, we utilized the loss and accuracy obtained by the models on the validation set to track their evolution during epochs, selecting the best-performing one once the training phase ended. Then, we evaluated such selected models on the external never-before-seen test set and reported their accuracy value. This way, we obtain a quantitative metric to compare the baseline architecture, the \textit{Cat} model, and our proposed \textit{Mulcat} architectures.

Ablation studies remove or damage specific components in a controlled setting to investigate all possible outcomes of system failure, thus understanding the contribution of a component to the overall system. The \textbf{ResNet18} (baseline) models already act as the ablation models for the remaining five types of models. Nevertheless, we wanted to do more than solely remove components. To gauge the significance of the values contained in causality maps and causality factors, we performed an additional test where we distort (i.e., damage) their information. We call these partially ablated versions of the networks \textbf{damaged}.
Concerning the \textbf{ResNet18 $+$ Cat} option, the only contribution of the causality map to the classification resides in the flattened elements that are concatenated to the actual (flattened) feature maps. Therefore, a natural \textit{damaged} network for such a setting would be to create a fictitious causality map filled with random probability values. We called this model the \textbf{ResNet18 $+$ Damaged-Cat}.
On the other hand, when it comes to the \textbf{ResNet18 $+$ Mulcat} option, the key functionality is to extract a vector of meaningful causality factors that serve as weights to the feature maps. Hence, we created the \textbf{ResNet18 $+$ Damaged-Mulcat} model, where we modify that vector to weigh features randomly rather than based on a principled way. This model comes in two variants according to the possible values of the causality factors mode, \textbf{m}. 
Indeed, when \textbf{m} $=$ \textit{full}, the $1 \times k $ vector of causality factors (i.e., weights) is replaced with a random vector of the same size with integer values ranging from $0$ (a feature map is never \textit{cause} of another feature) to $k-1$ (it is \textit{cause} of every other feature). Whereas, when \textbf{m} $=$ \textit{bool}, the values of the weights are randomly assigned to either $0$ or $1$. Since, in this setting, weights are hand-crafted, there is no need to consider the causality direction used; therefore, the \textit{damaged} study we performed is valid for both \textbf{d} $=$ \textit{causes} and \textbf{d} $=$ \textit{effects}.

To observe how the different architectures differ in terms of memory requirements, we track the size of the trained models (in megabytes) and the number of corresponding parameters (in millions). To compute the former, we don't want to rely on the file size of the saved models (e.g., \textit{.pth} files from PyTorch), as the file might be compressed. In fact, we calculate the number of parameters and buffers, multiply them by the element size, and accumulate these numbers.

\subsection{Qualitative evaluation}
To further investigate the possible benefits of integrating causality into CNNs for medical image classification, we performed explainable AI (XAI) experiments on the best-performing model for each type. Specifically, we aimed to obtain class activation maps (CAM) for the networks' decisions in all six types of models in our investigation: \textbf{Baseline} model, \textbf{Cat} model, \textbf{Mulcat-full-causes} model, \textbf{Mulcat-bool-causes} model, \textbf{Mulcat-full-effects} model, and \textbf{Mulcat-bool-effects} model. In all these models we assume ResNet18 as the backbone. Since investigating the variability of the visual output when changing the XAI method used is outside the scope of our work, we chose the popular Grad-CAM method \citep{selvaraju2017grad}, implemented in the \textit{pytorch-grad-cam} library \citep{jacobgilpytorchcam}. For the same reason, we selected the last convolutional layer of our architectures as the target layer for which we computed the CAM and performed the analysis with standard parameters. A more systematic analysis would require investigating the CAM output on all layers of the CNN and optimizing the smoothing parameters. 

To evaluate the quality and robustness of the produced CAMs, we followed the following criteria for the two datasets:
\begin{itemize}
    \item \textbf{BreakHis dataset}. To differentiate benign from malignant tumors, pathologists examine breast tissues at different magnification levels. Specifically, at $400\times$ magnification level, as the one used for our experiments, they analyze cytological features, such as shape and size of the nuclei, hyperchromatic nuclei, mitotic cells, and prominent nuclei \citep{young2013wheater}. To highlight cell nuclei, they employ Hematoxylin and Eosin stains, which make the nuclei appear dark purple or blue, while the other structures appear in shades of pink, red, and orange \citep{he2012histology}. For these reasons, we considered good explanations, the ones that focus on regions containing the dark purple/blue structures assumed as the nuclei of the cells.
     
    \item \textbf{PI-CAI dataset}. Based on the classification task, we considered explanations focusing on discriminative regions of the MRI (e.g., prostate gland area) to be better. In contrast, we considered explanations focusing on other structures, such as the rectum, bladder, or lateral muscle bundles, to be of lower quality and robustness.
\end{itemize}

\subsection{Additional experiments}
We conducted additional experiments on two very common application fronts to further test the effectiveness of our method. On the one hand, we proved that our module is easy to fit into existing convolutional models using other forms of visual attention, thus creating synergy. On the other, we verified its functioning in low-data scenarios, extending its applicability to Few-Shot Learning (FSL) \citep{fink2004object,fei2006one}.

\subsubsection{Integrating Bottleneck Attention Modules}
The bottleneck attention module (BAM) \citep{park2020simple,woo2018cbam} is a popular attention-based mechanism that, given a feature map, learns the attention map along two factorized axes, \textit{channel} and \textit{spatial}, to strengthen the representational power of CNNs. We thus investigated the addition of our module to models that already leveraged BAM. We used the same backbone as above (ResNet18) and placed multiple BAMs located after its layers $1$, $2$, and $3$, to build hierarchical attention. After training with the same strategy as the main study, we compared the performance of BAM-based regular models (\textbf{ResNet18 $+$ BAM}), BAM-based models integrating the \textit{Cat} method (\textbf{ResNet18 $+$ BAM $+$ Cat}), and BAM-based models that integrate our \textit{Mulcat} module (\textbf{ResNet18 $+$ BAM $+$ Mulcat}).

\subsubsection{Few-Shot experiments}
In addition to fully-supervised studies, in this paper, we extended our recent investigation into causality-driven one-shot learning (OSL) \citep{carloni2023causality} to the new BreakHis dataset, to understand how our \textit{Mulcat} methods worked under the shortage of annotated data in the medical imaging domain. To make the analyses consistent, we only considered the \textit{causes} direction in our \textit{Mulcat} models. We adopted the meta-learning strategy and formulated each task (i.e., episode) of the training process as an \textit{N-way 1-shot} classification problem, that is, to classify \textit{N} classes using only \textit{1} support image per class.

Regarding the PI-CAI dataset, we utilized a subset of the data containing lesions and the clinical question was tumor grading (i.e., predict aggressiveness). From a higher-level perspective than that of GS scores and ISUP groups, prostate lesions with GS $\leq3+3$ (ISUP $=1$) and with GS $ =3+4$ (ISUP $=2$) are considered low-grade (LG) tumors, while those with GS $>3+4$ (ISUP $>2$) are high-grade (HG) tumors. We considered only lesions whose GS was $\geq3+4$ (ISUP $\geq2$). As a result, we had eight classes of GS and four classes of ISUP in our dataset. The total number of images was $2049$ (from $382$ patients), which we divided into training (1611), validation ($200$), and testing ($238$) subsets, and resized to $128\times128$. We experimented with two classification scenarios on this dataset.
In the first scenario (\textbf{2-way}), the meta-training data are labeled to the four ISUP classes, and the model is meta-trained by randomly picking \textit{two} of the four classes in each task while distinguishing between LG and HG lesions during meta-testing. 
In the second scenario (\textbf{4-way}), we label meta-training data on the GS, and the model randomly picks \textit{four} of the eight GS classes in each task while distinguishing between four ISUP classes in meta-testing.

Regarding the BreakHis dataset, we used the same subsets as for the main study and considered two scenarios: in the \textbf{2-way} scenario, the meta-training is performed by randomly picking two of the eight classes of aggressiveness in each task, and meta-testing is done on the two high-level classes \textit{benign-vs-malignant}; in the \textbf{4-way} scenario, the meta-training is done on four out of eight random classes and the meta-testing is performed on four most prevalent classes (i.e., ductal carcinoma, fibroadenoma, phyllodes tumor, and tubular adenoma).

To increase the models' robustness to different data selections, we performed 600 meta-training tasks, 600 meta-validation tasks, and 600 meta-testing tasks for each experiment. To cope with the dataset unbalancing, we employed the AUC margin loss (AUCM) \citep{yang2022algorithmic} and the proximal epoch stochastic method (PESG) \citep{guo2020fast}, maximizing the Area Under the ROC curve (AUROC), which we used as our training and evaluation metric. Specifically, in \textit{2-way} experiments, we computed the binary AUROC, while we calculated the AUROC using the \textit{One-vs-rest} setting in \textit{4-way} experiments. Moreover, we evaluated the binary classification performance of the \textit{4-way} models by computing the AUROC of one significant class versus all the rest (i.e., $1$-vs-$3$): \textit{malignant} (ductal carcinoma) versus \textit{benign} (fibroadenoma, phyllodes tumor, tubular adenoma) for the BreakHis dataset, and \textit{LG} (ISUP=$2$) versus \textit{HG} (ISUP=$3$, ISUP=$4$, ISUP=$5$).
As with the main study, we performed ablation studies by repeating the OSL experiments with the \textit{damaged} version of our \textit{Mulcat} method.

\subsection{Implementation details}
All the experiments in this study ran on an NVIDIA A$100$ $40$ GB Tensor Core of the AI@Edge cluster of our Institute. We used Python $3.8.15$ and back-end libraries of PyTorch (version $1.13.0$, cuda $11.1$), together with other libraries such as scikit-learn $1.2.0$, grad-cam $1.4.8$, pydicom $2.3.1$, and pillow $9.4.0$. Docker version $20.10.11$ (build dea9396) was installed in the machine. To make results reproducible for each battery of experiments, we set a common seed for the random sequence generator of all the random processes and PyTorch functions. We release the codebase for our framework at \url{https://github.com/gianlucarloni/causality_conv_nets}.

\section{Results}
\label{sec:results}
\begin{table}[t]
    \begin{tabular*}{\tblwidth}{@{}LLLC@{}}
    \textbf{Architecture} & \textbf{Causality factors mode} & \textbf{Causality direction} & \textbf{Test set accuracy} [$\uparrow$] \\
    \hline
    \toprule
    \multicolumn{4}{c}{\textbf{\textit{BreakHis dataset (main study)}}}\\
    \midrule
    ResNet18 & - & - & $88.48_{0.77}$\\
    \hline
    ResNet18 $+$ Cat \citep{terziyan2023causality} & - & - & $85.77_{3.16}$\\
    \hline
    \multirow{4}{*}{ResNet18 $+$ \textbf{Mulcat (ours)}} 
     & Full & Causes & $91.32_{1.63}$ \\
    \cline{2-4}
     & Bool & Causes & $91.06_{1.32}$ \\
    \cline{2-4}
     & Full & Effects & $90.65_{1.14}$ \\
    \cline{2-4}
     & Bool & Effects & $91.59_{0.83}$ \\    
    \hline\\ 
    \multicolumn{4}{c}{\textbf{\textit{BreakHis dataset (ablation study)}}}\\
    \hline
    ResNet18 $+$ Damaged-Cat & - & - & $50.66_{1.93}$ \\
    \hline
    \multirow{2}{*}{ResNet18 $+$ Damaged-Mulcat} 
     & Full & Causes/Effects & $81.57_{0.50}$ \\
    \cline{2-4}
     & Bool & Causes/Effects &  $82.24_{2.15}$ \\    
    \bottomrule 

    \toprule
    \multicolumn{4}{c}{\textbf{\textit{PI-CAI dataset (main study)}}}\\
    \midrule
    ResNet18 & - & - & $68.38_{1.50}$ \\
    \hline
    ResNet18 $+$ Cat \citep{terziyan2023causality} & - & - & $70.07_{1.86}$ \\
    \hline
    \multirow{4}{*}{ResNet18 $+$ \textbf{Mulcat (ours)}} 
     & Full & Causes & $71.82_{1.00}$ \\
    \cline{2-4}
     & Bool & Causes & $70.31_{1.63}$ \\
    \cline{2-4}
     & Full & Effects & $69.96_{0.55}$ \\
    \cline{2-4}
     & Bool & Effects & $71.13_{0.36}$ \\
    \hline\\     
    \multicolumn{4}{c}{\textbf{\textit{PI-CAI dataset (ablation study)}}}\\
    \hline
    ResNet18 $+$ Damaged-Cat & - & - & $52.08_{0.93}$ \\
    \hline
    \multirow{2}{*}{ResNet18 $+$ Damaged-Mulcat} 
     & Full & Causes/Effects & $49.63_{0.88}$ \\
    \cline{2-4}
     & Bool & Causes/Effects &  $67.18_{0.99}$ \\    
    \bottomrule    
    \end{tabular*}
    \caption{Results of the main study and the ablation study for the best-performing models w.r.t the causality setting, the mode of computing the causality factors, and the direction used to encode the causality factors. We report accuracy results on the test set as the mean and standard deviation (in lower script) over four repetitions of the experiments with different seeds. The top half of the table refers to the BreakHis dataset, while the bottom half of it refers to the PI-CAI dataset.}
    \label{tab:best_results}
\end{table}

\subsection{Main study}
Table \ref{tab:best_results} shows the results of our main study for both datasets. We report the accuracy metric of the best-performing models on the external test set as the mean and standard deviation over four repetitions of the experiments with different seeds.
Regarding the \textbf{BreakHis dataset}, the baseline models (ResNet18) achieved an accuracy of $88.48$, while the competing method \textit{Cat} performed worse than the baseline, with an accuracy of $85.77$. On the other hand, our proposed \textbf{Mulcat} models, where the causality factors are ultimately computed in different ways depending on the mode \textit{m} and the direction \textit{d}, demonstrate higher performance than both previous choices. For instance, the \textit{full}-\textit{causes} models achieved $91.32$ accuracy, while their \textit{bool} version achieved an accuracy of $91.06$. On the other hand, the \textit{full}-\textit{effects} and \textit{bool}-\textit{effects} models reached accuracies of $90.65$ and $91.59$, respectively.
Regarding the \textbf{PI-CAI dataset}, while the baseline models achieved an accuracy of $68.38$, embedding causality in different forms improved performance. For instance, when the causality map was used with the \textit{Cat} version, the models achieved an accuracy of $70.07$. As for the \textbf{Mulcat} models, they all ranked above baseline, with the \textit{full}-\textit{causes} models that achieved $71.82$ accuracy and the \textit{bool} version achieving an accuracy of $70.31$. On the other hand, the \textit{full}-\textit{effects} models achieved $69.96$ accuracy, while using the \textit{bool} version led the models to reach an accuracy of $71.13$.

\subsection{Ablation study}
Table \ref{tab:best_results} also shows the results of purposely damaging the information contained in the causality maps and causality factors. These partial ablation studies for the two datasets reveal that, on BreakHis data, the \textbf{Damaged-Cat} models obtained an accuracy of $50.66$, and the \textbf{Damaged-Mulcat} models obtained accuracy values of $81.57$ and $82.24$ when using \textit{full} and \textit{bool} mode, respectively. As for the PI-CAI dataset, the \textbf{Damaged-Cat} models achieved an accuracy of $52.08$, and the \textbf{Damaged-Mulcat} models obtained accuracy values of $49.63$ and $67.18$ when using \textit{full} and \textit{bool} mode, respectively.     

\subsection{Memory requirements}
The size of the trained models and the number of corresponding parameters for both experiments are given in Table \ref{tab:memory_requirements}. While our \textbf{Mulcat} models increase memory demand by a negligible amount compared to their baseline counterparts ($+0.8\%$), using \textbf{Cat} models results in an overhead of up to approximately $+4.7\%$. 

\begin{table}
    \begin{tabular*}{\textwidth}{@{}LCC@{}}
    \textbf{Architecture} & \textbf{Model size (MB)} [$\downarrow$] & \textbf{Number of parameters ($\times 10^6$)} [$\downarrow$] \\
    \hline
    \toprule
    \multicolumn{3}{c}{\textit{\textbf{BreakHis dataset} (image size: $3\times128\times128$)}}\\
    \midrule
    ResNet18 & $42.73$ & $11.19$ \\
    \hline
    ResNet18 $+$ Cat \citep{terziyan2023causality} & $44.73$ & $11.72$ \\
    \hline
    ResNet18 $+$ \textbf{Mulcat (ours)} & $42.80$ & $11.21$ \\
    \hline
    
    \multicolumn{3}{c}{\textit{\textbf{PI-CAI dataset} (image size: $1\times96\times96$)}}\\
    \midrule
    ResNet18 & $42.68$ & $11.18$ \\
    \hline
    ResNet18 $+$ Cat \citep{terziyan2023causality} & $44.68$ & $11.70$ \\
    \hline
    ResNet18 $+$ \textbf{Mulcat (ours)} & $42.72$ & $11.19$ \\
    \hline
    \end{tabular*}
    \caption{Memory requirements in terms of model size (in megabytes) and number of parameters (in millions) of the ResNet18 baseline models, the ResNet18 backbone with the addition of the \textit{Cat} option, and the ResNet18 backbone with our \textit{Mulcat} module. Lower is better.}
    \label{tab:memory_requirements}
\end{table}

\subsection{XAI evaluations}
\begin{figure}
	\centering
		\includegraphics[width=0.6\textwidth]{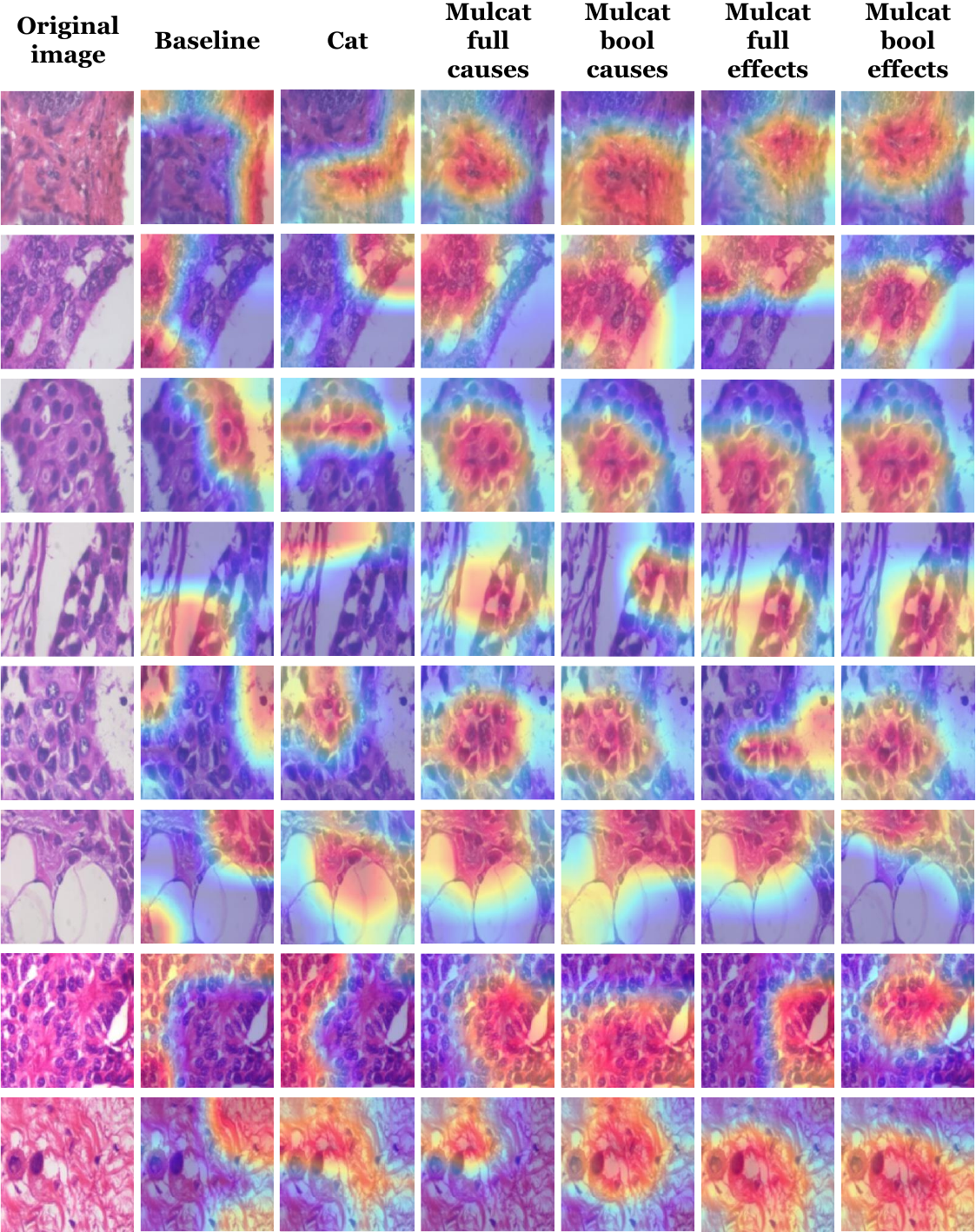}
	  \caption{Visual assessment of class activation maps for different malignant cases of the BreakHis dataset. Each row represents a different biopsy slide, and the columns represent the Grad-CAM outputs for the baseline model, the \textit{Cat} option, and all the proposed \textit{Mulcat} variants. ResNet18 is assumed as the backbone architecture in all models. Best seen in color.}
   \label{fig:CAM_malignant_breakhis}
\end{figure}

In addition to the quantitative experiments, we obtained qualitative results for the six models for each dataset by comparison of their CAMs given the same input test images. As an example, Figure \ref{fig:CAM_malignant_breakhis} shows the results for some BreakHis malignant cases for which all the models yielded the same correct prediction. Rows regard different bioptic slides, while columns represent from left to right the original input image, the CAM of the baseline (non-causality-driven) model, the CAM of the \textit{Cat} models, and the CAMs of our proposed \textit{Mulcat} models with their specific settings (i.e., direction \textit{d} and mode \textit{m}).

\begin{figure}
	\centering
		\includegraphics[width=0.6\textwidth]{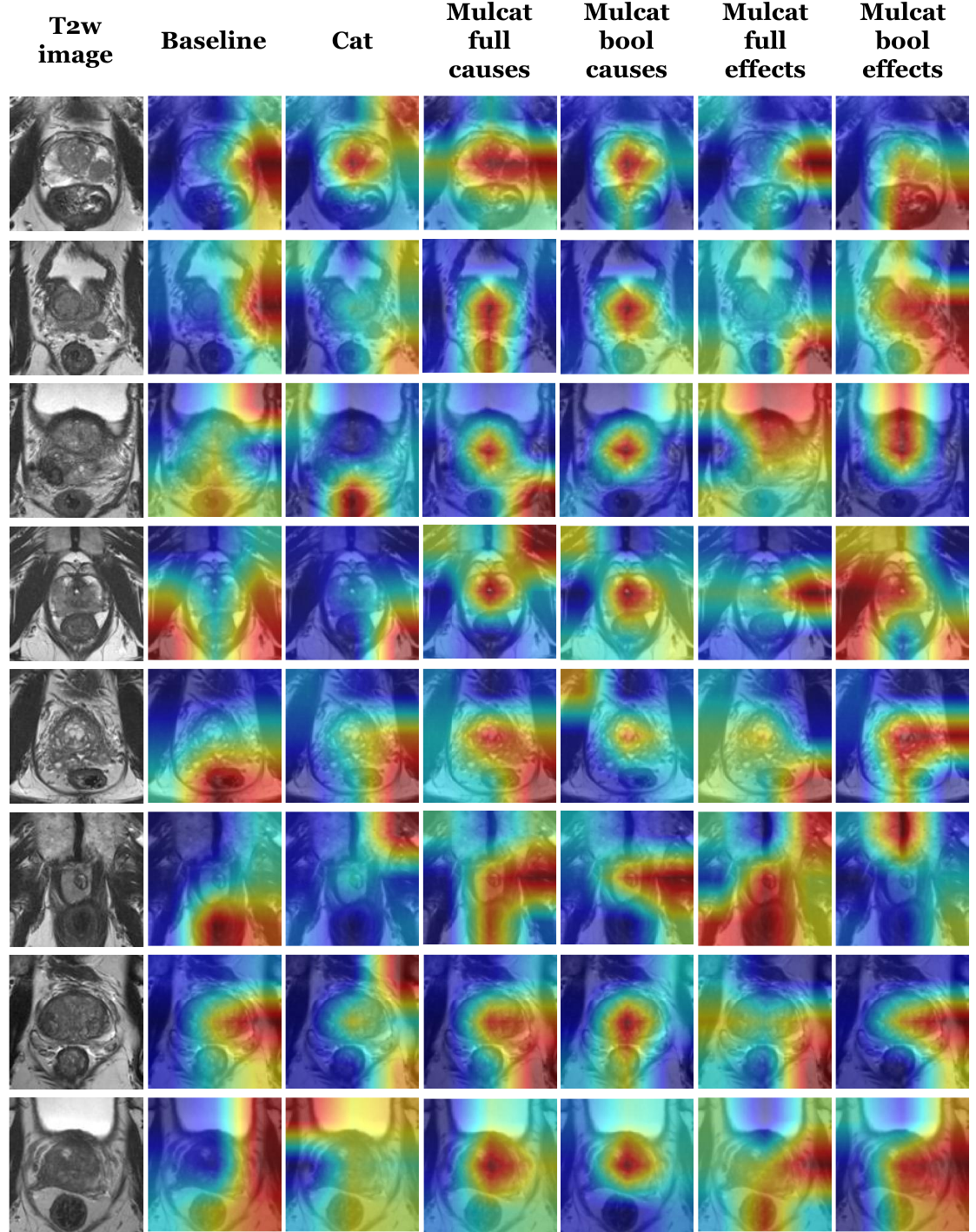}
	  \caption{Visual assessment of class activation maps for \textbf{cancerous} cases of the PI-CAI dataset. Each row represents a different scan, and the columns represent the Grad-CAM outputs for the baseline model, the \textit{Cat} option, and all the proposed \textit{Mulcat} variants. ResNet18 is assumed as the backbone architecture in all models. Best seen in color.}
   \label{fig:CAM_tumor}
\end{figure}

\begin{figure}
	\centering
		\includegraphics[width=0.6\textwidth]{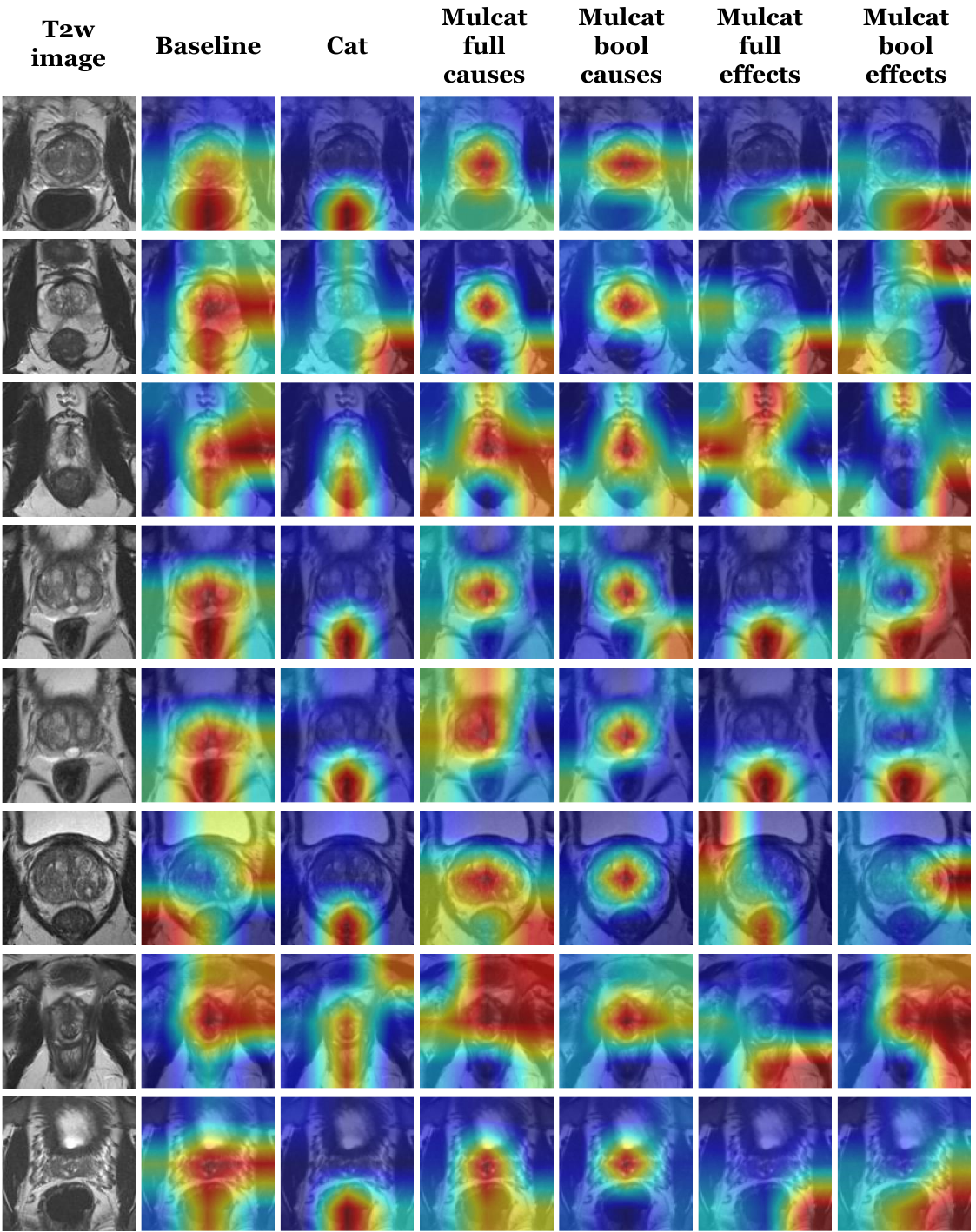}
	  \caption{Visual assessment of class activation maps for \textbf{no-tumor} cases of the PI-CAI dataset. Each row represents a different scan, and the columns represent the Grad-CAM outputs for the baseline model, the \textit{Cat} option, and all the proposed \textit{Mulcat} variants. ResNet18 is assumed as the backbone architecture in all models. Best seen in color.}
   \label{fig:CAM_notumor}
\end{figure}

Similarly, Figures \ref{fig:CAM_tumor} and \ref{fig:CAM_notumor} show results for the PI-CAI dataset on different cancerous and no-tumor cases, respectively. Again, rows represent several scans, while columns represent, from left to right, the original T2w input image and the CAMs of each configuration.

\subsection{Integration with BAMs}
In Table \ref{tab:best_results_BAM}, we show the result of integrating our \textit{Mulcat} module to convolutional networks that utilize BAM attention. Regarding the BreakHis dataset, the regular BAM-based models achieved an accuracy of $89.63$, and utilizing a \textit{Cat} option worsened performance ($85.77$). Conversely, almost all the models that integrated our \textit{Mulcat} modules performed better with a maximum accuracy of $92.07$. We obtained similar results for the PI-CAI dataset (max accuracy: $71.93$).

\begin{table}[t]
    \begin{tabular*}{\tblwidth}{@{}LLLC@{}}
    \textbf{Architecture} & \textbf{Causality factors mode} & \textbf{Causality direction} & \textbf{Test set accuracy} [$\uparrow$] \\
    \hline
    \toprule
    \multicolumn{4}{c}{\textbf{\textit{BreakHis dataset}}}\\
    \midrule   
    ResNet18 $+$ BAM \citep{park2020simple} & - & - & $89.63_{1.15}$\\
    \hline
    ResNet18 $+$ BAM $+$ Cat \citep{terziyan2023causality} & - & - & $85.77_{1.72}$\\
    \hline
    \multirow{4}{*}{ResNet18 $+$ BAM $+$ \textbf{Mulcat (ours)}} 
     & Full & Causes & $84.75_{2.01}$ \\
    \cline{2-4}
     & Bool & Causes & $90.65_{2.01}$ \\
    \cline{2-4}
     & Full & Effects & $91.87_{1.15}$ \\
    \cline{2-4}
     & Bool & Effects & $92.07_{0.86}$ \\   
    \bottomrule
    
    \toprule
    \multicolumn{4}{c}{\textbf{\textit{PI-CAI dataset}}}\\
    \midrule
    ResNet18 $+$ BAM \citep{park2020simple} & - & - & $68.00_{0.61}$\\
    \hline
    ResNet18 $+$ BAM $+$ Cat \citep{terziyan2023causality} & - & - & $66.09_{0.41}$\\
    \hline
    \multirow{4}{*}{ResNet18 $+$ BAM $+$ \textbf{Mulcat (ours)}} 
     & Full & Causes & $67.75_{1.99}$ \\
    \cline{2-4}
     & Bool & Causes & $71.25_{0.87}$ \\
    \cline{2-4}
     & Full & Effects & $71.93_{0.37}$ \\
    \cline{2-4}
     & Bool & Effects & $66.52_{1.67}$ \\    
    \bottomrule    
    \end{tabular*}
    \caption{The effect of integrating our \textit{Mulcat} method to models that utilize BAM modules, compared to baseline and competing models. We report accuracy results on the test set as the mean and standard deviation (in lower script) over four repetitions of the experiments with different seeds. The top half of the table refers to the BreakHis dataset, while the bottom half of it refers to the PI-CAI dataset.}
    \label{tab:best_results_BAM}
\end{table}

\subsection{One-Shot tasks}
The main results of our OSL analysis are reported in Table \ref{tab:best_results_OSL} for both datasets. We report all values as mean and standard deviation AUROC across all the $600$ meta-test tasks.
Regarding the BreakHis dataset, the \textit{2-way} experiment was where our \textit{Mulcat} module improved the models the most. Indeed, while the baseline achieved $0.51$ AUROC, we achieved up to $0.69$ AUROC with the ResNet18$+$\textit{Mulcat-Bool}. In contrast, we found this improvement to be lower in the case of \textit{4-way} experiments. Table \ref{tab:best_results_OSL} also shows the results for \textit{damaged Mulcat} models, which consistently performed worse than their \textit{Mulcat} counterparts.
Concerning the PI-CAI dataset, embedding our \textit{Mulcat} module improved the models in all scenarios, with a more pronounced improvement in the \textit{4-way 1-shot*}, where the models are trained to distinguish four classes (ISUP $2-5$), but the AUROC is computed between ISUP $2$ versus rest.

\begin{table}[t]
    \begin{tabular*}{0.925\textwidth}{@{}L|C|C|C|C@{}}
    \textbf{Architecture } & \textbf{Causality factors mode} & \textbf{2-way 1-shot} [$\uparrow$] & \textbf{4-way 1-shot} [$\uparrow$] & \textbf{4-way 1-shot*} [$\uparrow$]\\
    \hline
    \toprule
    \multicolumn{5}{c}{\textbf{\textit{BreakHis dataset (OSL study)}}}\\
    \midrule
    ResNet18 & - & $0.51_{0.15}$ & $0.59_{0.06}$ & $0.58_{0.20}$\\
    \hline
    \multirow{2}{*}{ResNet18 $+$ \textbf{Mulcat}} 
     & Full & $0.66_{0.24}$ & $0.59_{0.08}$ & $0.59_{0.15}$ \\
    \cline{2-5}
    & Bool & $0.69_{0.25}$ & $0.57_{0.09}$ & $0.57_{0.15}$ \\
    \hline     
    \multicolumn{5}{c}{\textbf{\textit{BreakHis dataset (OSL ablation study)}}}\\
    \hline
    \multirow{2}{*}{ResNet18 $+$ Damaged-Mulcat} 
     & Full & $0.51_{0.12}$ & $0.52_{0.06}$ & $0.52_{0.12}$ \\
    \cline{2-5}
     & Bool & $0.66_{0.25}$ & $0.50_{0.05}$ & $0.56_{0.13}$ \\ 
    \bottomrule
    
    \toprule
    \multicolumn{5}{c}{\textbf{\textit{PI-CAI dataset (OSL study)}}}\\
    \midrule
    ResNet18 & - & $0.54_{0.14}$ & $0.58_{0.07}$ & $0.59_{0.12}$\\
    \hline
    \multirow{2}{*}{ResNet18 $+$ \textbf{Mulcat}} 
     & Full & $0.55_{0.14}$ & $0.61_{0.07}$ & $0.71_{0.12}$ \\
    \cline{2-5}
    & Bool & $0.56_{0.14}$ & $0.61_{0.07}$ & $0.71_{0.12}$ \\
    \hline     
    \multicolumn{5}{c}{\textbf{\textit{PI-CAI dataset (OSL ablation study)}}}\\
    \hline
    \multirow{2}{*}{ResNet18 $+$ Damaged-Mulcat} 
     & Full & $0.53_{0.14}$ & $0.55_{0.06}$ & $0.55_{0.11}$ \\
    \cline{2-5}
     & Bool & $0.54_{0.14}$ & $0.57_{0.07}$ & $0.61_{0.12}$ \\    
    \bottomrule
    
    \end{tabular*}
    \caption{Results of the best-performing models under One-Shot Learning (OSL) settings in terms of AUROC values across all the 600 meta-test tasks as the mean and standard deviation (in lower script). The top half of the table refers to the BreakHis dataset, while the bottom half of it refers to the PI-CAI dataset. *: Trained to distinguish four classes, but the AUROC is computed with \textit{One-vs-rest}.}
    \label{tab:best_results_OSL}
\end{table}

\section{Discussion}\label{sec:discussion}
In this work, we presented a new method for automatically classifying medical images that use weak causal signals in the image to model how the presence of a feature in one part of the image affects the appearance of another feature in a different part of the image. Our plug-and-play \textit{Mulcat} module leverages causality maps in a new way and extracts multiplicative factors that eventually weight feature maps according to their causal influence in the scene. Our results seem to indicate that this lightweight, attention-inspired mechanism makes it possible to exploit weak causality signals in medical images to improve neural classifiers without any additional supervision signal.

In our main study, we assessed the effectiveness of our method under a fully-supervised learning scheme. In general, all the models obtained with our \textbf{Mulcat} implementation achieved higher performance than the \textit{baseline} (ResNet18) on the test set with both datasets (see Table \ref{tab:best_results}). This superiority ranged from a minimum of $+2.31$\% to a maximum of $+5.03$\%. On the other hand, utilizing the \textbf{Cat} option from \cite{terziyan2023causality} resulted in worse performance than most of our \textit{Mulcat} model and, with BreakHis data, even of the \textit{baseline}.
We found that most best-performing models used the \textit{Lehmer} method to get the causality map. Nevertheless, this choice comes with the drawback of necessitating more memory than the \textit{Max} method. We experimented with six different integer values for the parameter \textit{p} to sample the range of possible values. A possible improvement would be to let the network itself learn the parameter \textit{p} by back-propagation instead of giving it a fixed value beforehand.

To further confirm the numerical results of our studies, we conducted partially ablating studies on the actual influence of the causality factors on generating useful causality-driven feature maps. As anticipated, when we damage the causal weights by replacing them with random vectors, the accuracy of the final model is lower than its \textit{main study} counterpart (see Table \ref{tab:best_results}).
The \textit{Damaged Cat} performed worse than the \textit{None (baseline)} because the network was likely to be confused by the large number of random values that were concatenated to the actual extracted features. We expected that concatenating a random vector, not trained in back-propagation, would be worse than concatenating nothing at all.
In the \textit{Damaged Mulcat}, the weights multiplying the features maps are random, and, being untrained, they are re-computed at each iteration without any optimization from previous iterations. This results in scenarios where depending on the multiplication factors, irrelevant features are amplified while important ones are suppressed. When \textit{Damaged Mulcat} is used with the \textit{full} option, this behavior is more pronounced (random weights can have very high values, up to $k-1$), and performance is low (even lower than baseline for PI-CAI data) because the network assigns a lot of importance to these potentially incorrect features. In contrast, when the \textit{bool} option is used, this behavior is mitigated (random weights have a maximum value of $1$), so the degree of confusion of the network is reduced, and performance is higher.
Experiencing reduced performance when the causality maps and factors are completely ablated or partially corrupted suggests that our module is computing something significant. This observation indicates that, even if weak, the causality signals learned during training assist the network to perform better.

Although we notice the improvement of our \textbf{Mulcat} models over the baseline from a quantitative point of view, it seems that the different combinations of mode \textbf{m} and direction \textbf{d} behaved roughly the same way.
Thus, we wanted to investigate the potential benefits of our proposition on a different level. We deepened the analysis and found that significant differences can emerge on the XAI side, supporting the role of causality in explainability. 
On BreakHis data, the \textbf{Mulcat-full} and \textbf{Mulcat-bool} models manage to focus on regions densely populated by nuclei, which is what pathologists do at this magnification level. Instead, \textit{Baseline} models often pay attention to lateral zones or regions less critical for the malignancy classification. The latter behavior is also observed in the \textit{Cat} models, which frequently focus on lateral, small, and/or irrelevant portions for classification purposes (see Figure \ref{fig:CAM_malignant_breakhis}). 
As for the PI-CAI dataset, where the field of view is larger and comprises many different anatomic structures other than the prostate, we noticed a trend that \textbf{Mulcat-full-causes} and \textbf{Mulcat-bool-causes} are consistently more focused on the discriminative parts of the image (e.g., prostate gland area). Conversely, the other options led to models that often looked at the rectum, bladder, or lateral muscle bundles (see Figures \ref{fig:CAM_tumor} and \ref{fig:CAM_notumor}). This fact confirms our hypothesis that using causes and not effects allows the network to obtain more faithful results.

Among the methods that exploit the information of causality maps, \textbf{Cat} proves to be one of the worst. That is evident both quantitatively and qualitatively. The reason for this behavior could be the considerable complexity added to the model to account for all the combinations of feature maps. In fact, on the classifier, the number of input neurons goes from $n\times n\times k$ to $n\times n\times k + k\times k$, which with high $k$ results in thousands of additional connections (e.g., $512\times 512=262144$ new neurons for a ResNet18).
The overhead induced by the \textit{Cat} method is quantitatively confirmed by the memory requirements summarized in Table \ref{tab:memory_requirements}. Instead, our \textit{Mulcat} method increases the memory demand by a negligible amount compared to the baseline, promoting it as a low-cost improvement of regular architectures.

We conducted additional experiments to further add evidence of improved performance through our method. First, we showed that our \textit{Mulcat} module can be easily integrated into existing architectures, such as attention-based BAM networks, and can create synergy in improving performance (see Table \ref{tab:best_results_BAM}). Indeed, utilizing \textit{Mulcat} led to an increase of up to $+2.72\%$ over the regular BAM-ResNet18 and of $+7.34\%$ over the BAM-ResNet18 that used \textit{Cat}, for the BreakHis dataset. Similarly, BAM-ResNet18 which utilized our \textit{Mulcat} option on the PI-CAI dataset achieved up to $+5.78\%$ and $+8.83\%$ w.r.t regular BAM-ResNet18 and \textit{Cat} BAM-ResNet18, respectively.

Second, to tend towards a more generalized demonstration of classification problems, it was interesting to understand how our \textit{Mulcat} method worked in practical application situations, such as the shortage of annotated data in the medical imaging domain. Thus, we performed One-Shot Learning (OSL) experiments, both in \textit{2-way} and \textit{4-way} settings (see Table \ref{tab:best_results_OSL}). Our findings suggest that using \textit{Mulcat} can be an effective choice even in low-data scenarios. Indeed, when performing binary classification over the BreakHis data (i.e., \textit{benign-vs-malignant}), ResNet18$+$Mulcat achieved an accuracy up to $+35.2\%$ compared to the baseline, while this improvement was broadly reduced when performing \textit{4-way} experiments. On the other hand, it is on the \textit{4-way} scenarios with PI-CAI data that ResNet18$+$Mulcat outperformed the baseline the most, with an increase of $20.3\%$ accuracy for the \textit{4-way 1-shot*} setting. 

One of the limitations of our work is that we used only ResNet18 as the backbone architecture to extract latent representations for the different implementations, although this is consistent with the pilot nature of our study. 
Moreover, we acknowledge that our methods consider potential causal relationships in pairs rather than among more than two features. That, of course, can lead to suboptimal results, given the impossibility of excluding confounders. In future experiments, we would be interested in extending the operation to more variables and devising variations inspired by the classic PC algorithms of the literature on causal discovery in tabular data \citep{spirtes1991algorithm}.

There could be other directions to explore from our work, both on the application and architectural level. It would be interesting to draw inspiration from the multi-depth, visual attention \citep{jetley2018learn,yan2019melanoma,schlemper2019attention}, which extracts information from the convolutional encoder at different depths (local and global features). In this regard, one could extract the causality map from the internal layers of the network (not only from the last one).
Our work could also be expanded by proposing new methods to combine causal information besides concatenation and weighting of feature maps and by experimenting with ensemble methods.
Additionally, visualizations such as that in Figure \ref{fig_cmapvisual} suggest one could potentially conceive a low-cost self-supervised feature pruning based on similarities of features across rows and columns of causality maps. That could help to disregard redundant features and consequently lower model complexity in a data-driven way. In the end, we foresee a possible integration of our methods within the convolutional block of generative models such as GANs \citep{goodfellow2020generative} and diffusion models \citep{ho2020denoising,rombach2022high}, to guide the generation of more realistic images.

\section{Conclusions}
\label{sec:conclusions}
In this work, we introduced a novel technique to discover and exploit weak causal signals directly from medical images via neural networks for classification purposes. Our method consists of a CNN backbone and a causality-factors extractor module, which computes weights for the feature maps to enhance each feature map according to its causal influence in the image's scene in an attention-inspired fashion. We developed different architecture variants and empirically evaluated all of our models on two public datasets of medical images for cancer diagnosis. Moreover, we verified that our module can create synergies when introduced in existing attention-based architectures, and we verified its applicability to few-shot learning settings. 

Our findings demonstrate how minor modifications to traditional models can enhance them. Indeed, our lightweight module can be easily integrated into regular CNN classification systems and produce better models without requiring additional trainable parameters. It enhances the overall classification results and makes the model focus more precisely on the critical regions of the image, leading to more accurate and robust predictions.
This aspect is crucial in medical imaging, where accurate and reliable classification is essential for effective diagnosis and treatment planning. Nevertheless, what we propose in this paper may have a broader significance, such as non-medical tasks or application to other data types, such as videos.
We believe that the new elements we introduce with our work are a way to connect machine vision and causal reasoning in a novel way, especially when no prior knowledge of the data is available, adding a unique dimension to the framework. 

\section{Acknowledgements}
The research leading to these results has received funding from the European Union’s Horizon 2020 research and innovation program under grant agreement No 952159 (ProCAncer-I), and partially from the Regional Project PAR FAS Tuscany - NAVIGATOR. The funders had no role in the design of the study, collection, analysis, and interpretation of data, or writing the manuscript.

\section{Competing interests statement}
The authors declare no competing interests.

\printcredits

\bibliographystyle{apalike}

\bibliography{cas-sc-main}

\bio{}
\endbio

\endbio

\end{document}